\documentclass[11pt,a4paper]{article}
\usepackage[hyperref]{emnlp2018}
\usepackage{latexsym}
\usepackage{times}
\usepackage{url}
\usepackage{latexsym}
\usepackage{mathtools}
\usepackage{amsthm}
\usepackage{color}
\usepackage{multirow}
\usepackage{tablefootnote}
\usepackage{amssymb}
\usepackage{pifont}
\usepackage{epstopdf}
\aclfinalcopy 

\renewcommand{\vec}[1]{\mathbf{#1}}

\DeclareMathOperator*{\argmax}{arg\,max}

\newcommand{\myterm}{{\bullet}}

\title{Constituent Parsing as Sequence Labeling}

\author{Carlos G\'{o}mez-Rodr\'{i}guez \\
  Universidade da Coru\~{n}a \\
  FASTPARSE Lab, LyS Group \\
  Departamento de Computaci\'{o}n \\
  Campus de Elvi\~{n}a s/n, 15071 \\ A Coru\~{n}a, Spain \\
  {\tt carlos.gomez@udc.es} \\
  \\\And
  David Vilares\\
  Universidade da Coru\~{n}a \\
  FASTPARSE Lab, LyS Group \\
  Departamento de Computaci\'{o}n \\
  Campus de Elvi\~{n}a s/n, 15071 \\ A Coru\~{n}a, Spain \\
  {\tt david.vilares@udc.es} \\}

\date{}
\begin{document}
\maketitle


\vspace{0.2cm}
\begin{abstract}

We introduce a method to reduce constituent parsing to sequence labeling. For each word $w_t$, it generates a label that encodes: (1) the number of ancestors in the tree that the words $w_t$ and $w_{t+1}$ have in common, and (2) the nonterminal symbol at the lowest common ancestor. We first prove that the proposed encoding function is injective for any tree without unary branches. In practice, the approach is made extensible to all constituency trees by collapsing unary branches. We then use the \textsc{ptb} and \textsc{ctb} treebanks as testbeds and propose a set of fast baselines. We achieve 
90.7\% F-score
on the \textsc{ptb} test set, outperforming the \newcite{vinyals2015grammar} sequence-to-sequence parser. In addition, sacrificing some accuracy, our approach achieves the fastest constituent parsing speeds reported to date on \textsc{ptb} by a wide margin.\footnote{This is a revision  with improved results of our paper originally published in EMNLP 2018. The previous version contained a bug where the script \textsc{evalb} was not considering the COLLINS.prm parameter file.}
\end{abstract}

\section{Introduction}
Constituent parsing is a core problem in \textsc{nlp} where the goal is to obtain the syntactic structure of sentences expressed as a phrase structure tree.

Traditionally, constituent-based parsers have been built relying on chart-based, statistical models \cite{collins1997three,charniak2000maximum,petrov2006learning}, which are accurate but slow, with typical speeds well below 10 sentences per second on modern CPUs \cite{kummerfeldCorral}.

Several authors have proposed more efficient approaches which are helpful to gain speed while preserving (or even improving) accuracy. \newcite{sagae2005classifier} present a classifier for constituency parsing that runs in linear time by relying on a shift-reduce stack-based algorithm, instead of a grammar. It is essentially an extension of transition-based dependency parsing \cite{nivre2003efficient}. This line of research has been polished through the years \cite{Wang2006,zhu2013fast,DyerRecurrent2016,Liu2017InOrder,Fer2018Faster}.

With an aim more related to our work, other authors have reduced constituency parsing to tasks that can be solved faster or in a more generic way. \newcite{Fer2015Parsing} reduce phrase structure parsing to dependency parsing.
They  propose an intermediate representation where dependency labels from a head to its dependents encode the nonterminal symbol and an attachment order that is used to arrange nodes into constituents. 
Their approach makes it possible to use off-the-shelf dependency parsers for constituency parsing.
In a different line, \newcite{vinyals2015grammar} address the problem by relying on a sequence-to-sequence model where trees are linearized in a depth-first traversal order. Their solution can be seen as a machine translation model that maps a sequence of words into a parenthesized version of the tree. \newcite{ChoeChar2016} recast parsing as language modeling. They train a generative parser that obtains the phrasal structure of sentences by relying on the \newcite{vinyals2015grammar} intuition and on the \newcite{zaremba2014recurrent} model to build the basic language modeling architecture. 

More recently, \newcite{ShenDistance2018} propose an architecture to speed up the current state-of-the-art chart parsers trained with deep neural networks \cite{stern2017minimal,Kitaev2018Constituency}. They introduce the concept of \emph{syntactic distances}, which specify the order in which the splitting points of a sentence will be selected. The model learns to predict such distances, to then recursively partition the input in a top-down fashion.

\paragraph{Contribution} We propose a method to transform constituent parsing into sequence labeling. This reduces it to the complexity of tasks such as part-of-speech (PoS) tagging, chunking or named-entity recognition. The contribution is two-fold. 

First, we describe a method to linearize a tree into a sequence of labels (\S \ref{section-encoding}) of the same length of the sentence minus one.\footnote{A last dummy label is generated to fulfill the properties of sequence labeling tasks.} The label generated for each word encodes the 
number of common ancestors in the constituent tree between that word and the next, and 
the nonterminal symbol associated with the lowest common ancestor.
We prove that the encoding function is injective for any tree without unary branchings. After applying collapsing techniques,  the method can parse unary chains.

Second, we use such encoding to present different baselines that can effectively predict the structure of sentences (\S \ref{section-seq2seq}). To do so, we rely on a recurrent sequence labeling model based on  \textsc{bilstm}'s \cite{hochreiter1997long,yang2017ncrf}. We also test other models inspired in classic approaches for other tagging tasks \cite{Schmid1994Part,sha2003shallow}. We use the Penn Treebank (\textsc{ptb}) and the Penn Chinese Treebank (\textsc{ctb}) as testbeds.

The comparison against \newcite{vinyals2015grammar}, the closest work to ours, shows that our method is able to train more accurate parsers. This is in spite of the fact that our approach addresses constituent parsing as a sequence labeling problem, which is simpler than a sequence-to-sequence problem, where the output sequence has variable/unknown length.
Despite being the first sequence labeling method for constituent parsing, our baselines achieve decent accuracy results in comparison to models coming from mature lines of research, and their speeds are the fastest reported to our knowledge.

\section{Linearization of \emph{n}-ary trees}\label{section-encoding}

\paragraph{Notation and Preliminaries} 
In what follows, we use bold style to refer to vectors and matrices (e.g $\vec{x}$ and $\vec{W}$).
Let $\vec{w}$=$[w_{1},w_{2},...,w_{|w|}]$ be an input sequence of words, where $w_{i} \in V$. Let $T_{|w|}$ be the set of constituent trees with $|w|$ leaf nodes that have no unary branches. For now, we will assume that the constituent parsing problem consists in mapping each sentence $\vec{w}$ to a tree in $T_{|w|}$, i.e., we assume that correct parses have no unary branches. We will deal with unary branches later.

To reduce the problem to a sequence labeling task, we define 
a set of labels $L$ 
that allows us to encode each tree in $T_{|w|}$ as a unique sequence of labels in $L^{(|w|-1)}$, via an encoding function $\Phi_{|w|}: T_{|w|} \rightarrow L^{(|w|-1)}$.
Then, we can reduce the constituent parsing problem to a sequence labeling task where the goal is to predict a function $F_{|w|,\theta}: V^{|w|} \rightarrow L^{|w|-1}$, where $\theta$ are the parameters to be learned. To parse a sentence, we label it and then decode the resulting label sequence into a constituent tree, i.e., we apply $F_{|w|,\theta} \circ \Phi_{|w|}^{-1}$.

For the method to be correct, we need the encoding of trees to be complete (every tree in $T_{|w|}$ must be expressible as a label sequence, i.e., $\Phi_{|w|}$ must be a function, so we have full coverage of constituent trees) and injective (so that the inverse function $\Phi_{|w|}^{-1}$ is well-defined). Surjectivity is also desirable, so that the inverse is a function on $L^{|w|-1}$, and the parser outputs a tree for any sequence of labels that the classifier can generate. 

We now define our $\Phi_{|w|}$ and show that it is total and injective. Our encoding is not surjective per se. We handle ill-formed label sequences in \S \ref{section-limitations}.

\subsection{The Encoding} 
\label{sec:encoding}
Let $w_i$ be a word located at position $i$ in the sentence, for $1 \le i \le |w|-1$. We will assign it a 2-tuple label $l_i = (n_i,c_i)$, where:
$n_i$ is an integer that encodes the number of common ancestors between $w_i$ and $w_{i+1}$, 
and $c_i$ is the nonterminal symbol at the lowest common ancestor.

\paragraph{Basic encodings}

\begin{figure}[hbtp]
\centering
\includegraphics[width=1\columnwidth]{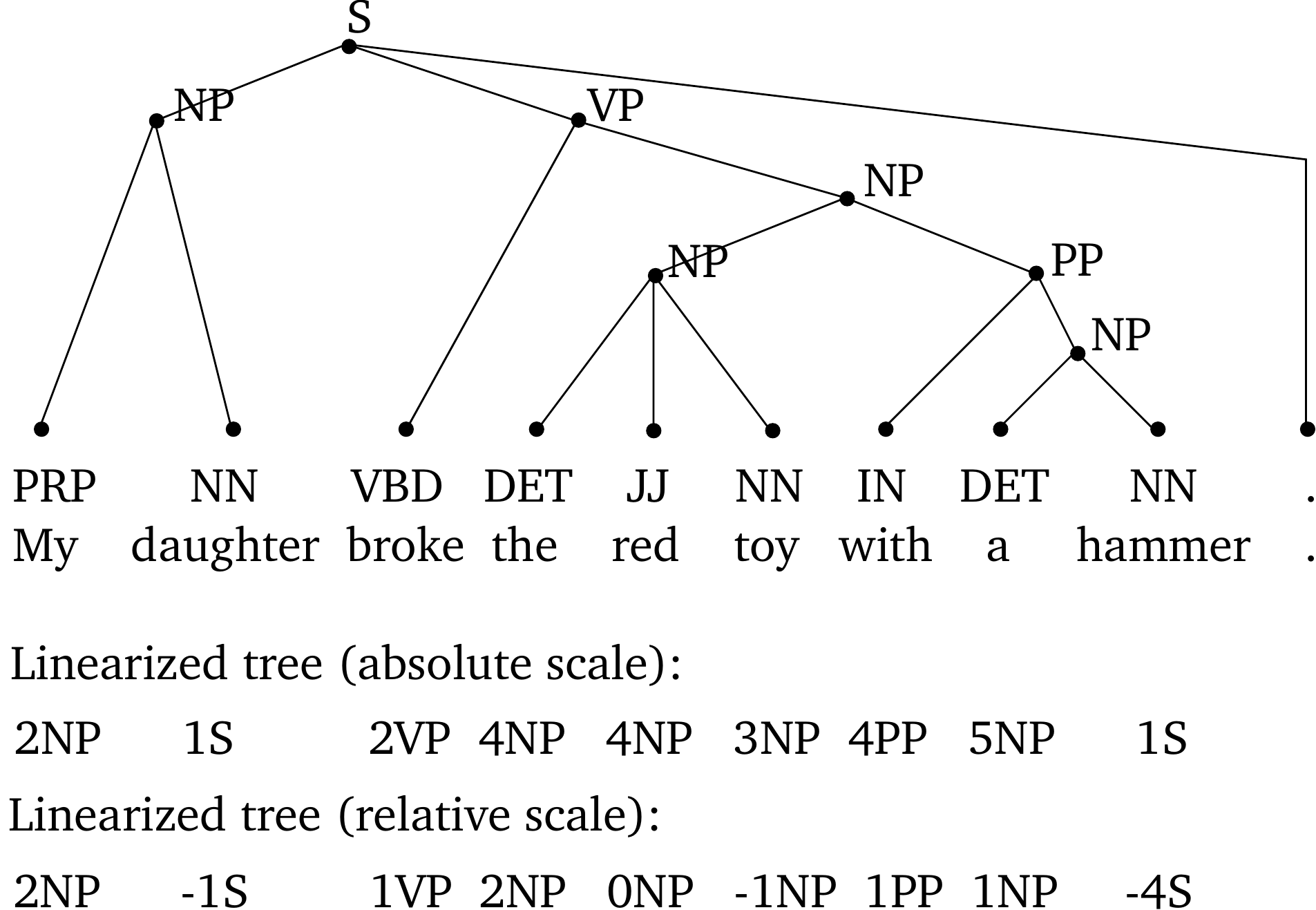}
\caption{\label{f-running-example} An example of a constituency tree linearized applying both absolute and relative scales.}
\end{figure}

The number of common ancestors may be encoded in several ways. 
\begin{enumerate}
\item Absolute scale: The simplest encoding is to make $n_i$ directly equal to the number of ancestors in common between $w_i$ and $w_{i+1}$. 
\item Relative scale: A second and better variant consists in making $n_i$ represent the difference with respect to the number of ancestors encoded in $n_{i-1}$.
Its main advantage is that the size of the label set is reduced considerably.
\end{enumerate}
Figure \ref{f-running-example} shows an example of a tree linearized according to both absolute and relative scales.

\paragraph{Encoding for trees with exactly $k$ children}

\begin{figure}[hbtp]
\centering
\includegraphics[width=1\columnwidth]{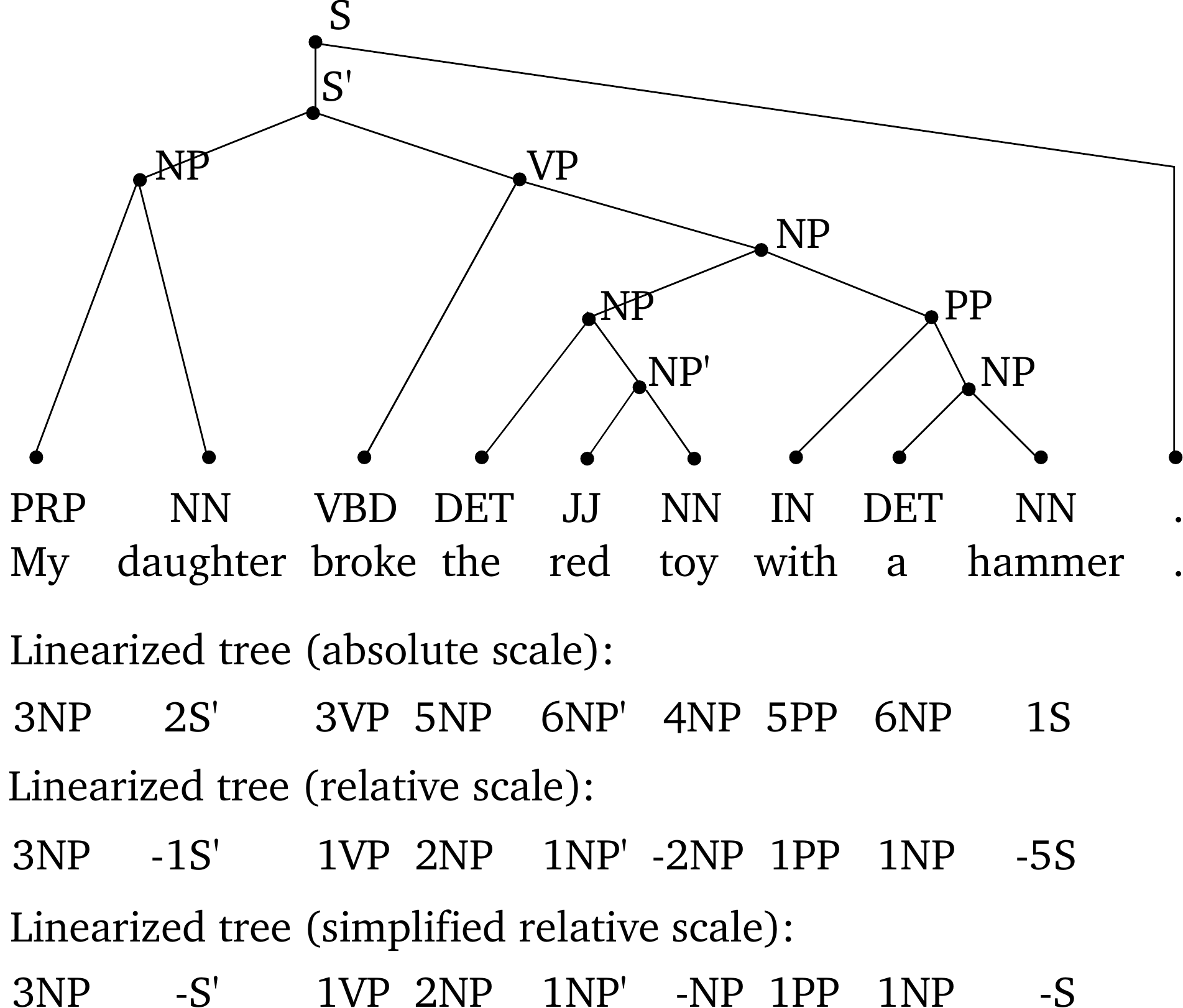}
\caption{\label{f-running-example-binarized} An example of a binarized constituency tree, linearized both applying absolute and relative scales.}
\end{figure}

 For trees where all branchings have exactly $k$ children, it is possible to obtain a even more efficient linearization in terms of number of labels. To do so, we take the relative scale encoding as our starting point. If we build the tree incrementally in a left-to-right manner from the labels, if we find a negative $n_i$, we will need to attach the word $w_{i+1}$ (or a new subtree with that word as its leftmost leaf) to the $(-n_i+2)$th node in the path going from $w_i$ to the root. If every node must have exactly $k$ children, there is only one valid negative value of $n_i$: the one pointing to the first node in said path that has not received its $k$th child yet. Any smaller value would leave this node without enough children (which cannot be fixed later due to the left-to-right order in which we build the tree), and any larger value would create a node with too many children.
Thus, we can map negative values to a single label.
Figure \ref{f-running-example-binarized} shows an example for the case of binarized trees ($k=2$).

\paragraph{Links to root} Another variant  emerged from the empirical observation that some tokens that are usually linked to the root node (such as the final punctuation in Figure \ref{f-running-example}) were particularly difficult to learn for the simpler baselines. To successfully deal with these cases in practice, it makes sense to consider a simplified annotation scheme where a node is assigned a special tag (\textsc{root}, $c_i$) when it is directly linked to the root of the tree.\\
 
From now on, unless otherwise specified, we use the relative scale without the simplification for exactly $k$ children. This will be the encoding used in the experiments (\S \ref{section-experiments}),
because the size of the label set is significantly lower than the one obtained by relying on the absolute one. Also, it works directly with non-binarized trees, in contrast to the encoding that we introduce for trees with exactly $k$ children, which is described only for completeness and possible interest for future work. For the experiments (\S \ref{section-experiments}), we also use the special tag (\textsc{root}, $c_i$) to further reduce the size of the label set and to simplify the classification of tokens connected to the root, where $|n_i|$ is expected to be large.

\subsection{Theoretical correctness}

We now prove that $\Phi_{|w|}$ is a total function and injective for any tree in $T_{|w|}$. We remind that trees in this set have no unary branches. Later (in \S \ref{section-limitations}) we describe how we deal with unary branches. To prove correctness, we use the relative scale. Correctness for the other scales follows trivially.

\paragraph{Completeness} 
Every pair of nodes in a rooted tree has at least one common ancestor, and a unique lowest common ancestor. Hence, for any tree in $T_{|w|}$, the label $l_i = (n_i,c_i)$ defined in Section \ref{sec:encoding} is well-defined and unique for each word $w_i$, $1 \le i \le |w|-1$; and thus $\Phi_{|w|}$ is a total function from $T_{|w|}$ to $L^{(|w|-1)}$.

\paragraph{Injectivity} 
The encoding method must ensure that any given sequence of labels corresponds to exactly one tree. Otherwise, we have to deal with ambiguity, which is not desirable. 

For simplicity, we will prove injectivity in two steps. First, we will show that the encoding is injective if we ignore nonterminals (i.e., equivalently, that the encoding is injective for the set of trees resulting from replacing all the nonterminals in trees in $T_{|w|}$ with a generic nonterminal $X$). Then, we will show that it remains injective when we take nonterminals into account.

For the first part, let $\tau \in T_{|w|}$ be a tree where nonterminals take a generic value $X$.
We represent the label of the $i$th leaf node as $\myterm_i$. Consider the representation of $\tau$ as a bracketed string, where a single-node tree with a node labeled $A$ is represented by $(A)$, and a tree rooted at $R$ with child subtrees $C_1 \ldots C_n$ is represented as $(R(C_1 \ldots C_n))$. 

Each leaf node will appear in this string as a substring $(\myterm_i)$. Thus, the parenthesized string has the form $\alpha_0 (\myterm_1) \alpha_1 (\myterm_2) \ldots \alpha_{|w|-1}(\myterm_{|w|})\alpha_w$, where the $\alpha_i$s are strings that can only contain brackets and nonterminals, as by construction there can be no leaf nodes between $(\myterm_i)$ and $(\myterm_{i+1})$. 

We now observe some properties of this parenthesized string. First, note that each of the substrings $\alpha_i$ must necessarily be composed of zero or more closing parentheses followed by zero or more opening parentheses with their corresponding nonterminal, i.e., it must be of the form $[)]^*[(X]^*$. This is because an opening parenthesis followed by a closing parenthesis would represent a leaf node, and there are no leaf nodes between $(\myterm_i)$ and $(\myterm_{i+1})$ in the tree.

Thus, we can write $\alpha_i$ as $\alpha_{i)} \alpha_{i(}$, where $\alpha_{i)}$ is a string matching the expression $[)]^*$ and $\alpha_{i(}$ a string matching the expression $[(X]^*$. With this, we can write the parenthesized string for $\tau$ as 
\[
\alpha_{0)}\alpha_{0(} (\myterm_1) \alpha_{1)}\alpha_{1(} (\myterm_2) 
\alpha_{2)}\alpha_{2(}
\ldots 
(\myterm_{|w|})\alpha_{|w|)}\alpha_{|w|(}.
\]
Let us now denote by $\beta_i$ the string $\alpha_{i-1(} (\myterm_i) \alpha_{i)}$. Then, and taking into account that $\alpha_{0)}$ and $\alpha_{w(}$ are trivially empty in the previous expression due to bracket balancing, the expression for the tree becomes simply $\beta_1 \beta_2 \ldots \beta_{|w|}$, where we know, by construction, that each $\beta_i$ is of the form $[(X]^* (\myterm_i) [)]^*$. 

Since we have shown that each tree in $T_{|w|}$ uniquely corresponds to a string $\beta_1 \beta_2 \ldots \beta_{|w|}$, to show injectivity of the encoding, it suffices to show that different values for a $\beta_i$ generate different label sequences.

To show this, we can say more about the form of $\beta_i$: it must be either of the form $[(X]^* (\myterm_i)$ or of the form $(\myterm_i) [)]^*$, i.e., it is not possible that $\beta_i$ contains both opening parenthesis before the leaf node and closing parentheses after the leaf node. This could only happen if the tree had a subtree of the form $(X(\myterm_i))$, but this is not possible since we are forbidding unary branches.

Hence, we can identify each $\beta_i$ with an integer number $\delta(\beta_i)$: $0$ if $\beta_i$ has neither opening nor closing parentheses outside the leaf node, $+k$ if it has $k$ opening parentheses, and $-k$ if it has $k$ closing parentheses. It is easy to see that $\delta(\beta_1)\delta(\beta_2)\ldots\delta(\beta_{|w|-1})$ corresponds to the values $n_i$ in the relative-scale label encoding of the tree $\tau$. To see this, note that the number of unclosed parentheses at the point right after $\beta_i$ in the string exactly corresponds to the number of common ancestors between the $i$th and $(i+1)$th leaf nodes. A positive $\delta(\beta_i) = k$ corresponds to opening $k$ parentheses before $\beta_i$, so the number of common ancestors of $w_i$ and $w_{i+1}$ will be $k$ more than that of $w_{i-1}$ and $w_i$. A negative $\delta(\beta_i) = -k$ corresponds to closing $k$ parentheses after $\beta_i$, so the number of common ancestors will conversely decrease by $k$. A value of zero means no opening or closing parentheses, and no change in the number of common ancestors.

Thus, different parenthesized strings $\beta_1 \beta_2 \ldots \beta_{|w|}$ generate different label sequences, which proves injectivity ignoring nonterminals (note that $\delta(\beta_{|w|})$ does not affect injectivity as it is uniquely determined by the other values: it corresponds to closing all the parentheses that remain unclosed at that point).

It remains to show that injectivity still holds when nonterminals are taken into account. Since we have already proven that trees with different structure produce different values of $n_i$ in the labels, it suffices to show that trees with the same structure, but different nonterminals, produce different values of $c_i$. Essentially, this reduces to showing that every nonterminal in the tree is mapped into a concrete $c_i$.  
That said, consider a tree $\tau \in T_{|w|}$, and some nonterminal $X$ in $\tau$. Since trees in $T_{w}$ do not have unary branches, $X$ has at least two children. Consider the rightmost word in the first child subtree, and call it $w_i$. Then, $w_{i+1}$ is the leftmost word in the second child subtree, and $X$ is the lowest common ancestor of $w_i$ and $w_{i+1}$. Thus, $c_i = X$, and a tree with identical structure but a different nonterminal at that position will generate a label sequence with a different value of $c_i$. This concludes the proof of injectivity.

\subsection{Limitations}\label{section-limitations}

We have shown that our proposed encoding is a total, injective function  from trees without unary branches with yield of length $|w|$ to sequences of $|w|-1$ labels. This will serve as the basis for our reduction of constituent parsing to sequence labeling. However, to go from theory to practice, we need to overcome two limitations of the theoretical encoding: non-surjectivity and the inability to encode unary branches. Fortunately, both can be overcome with simple techniques.

\begin{figure}[hbtp]
\centering
\includegraphics[width=0.7\columnwidth]{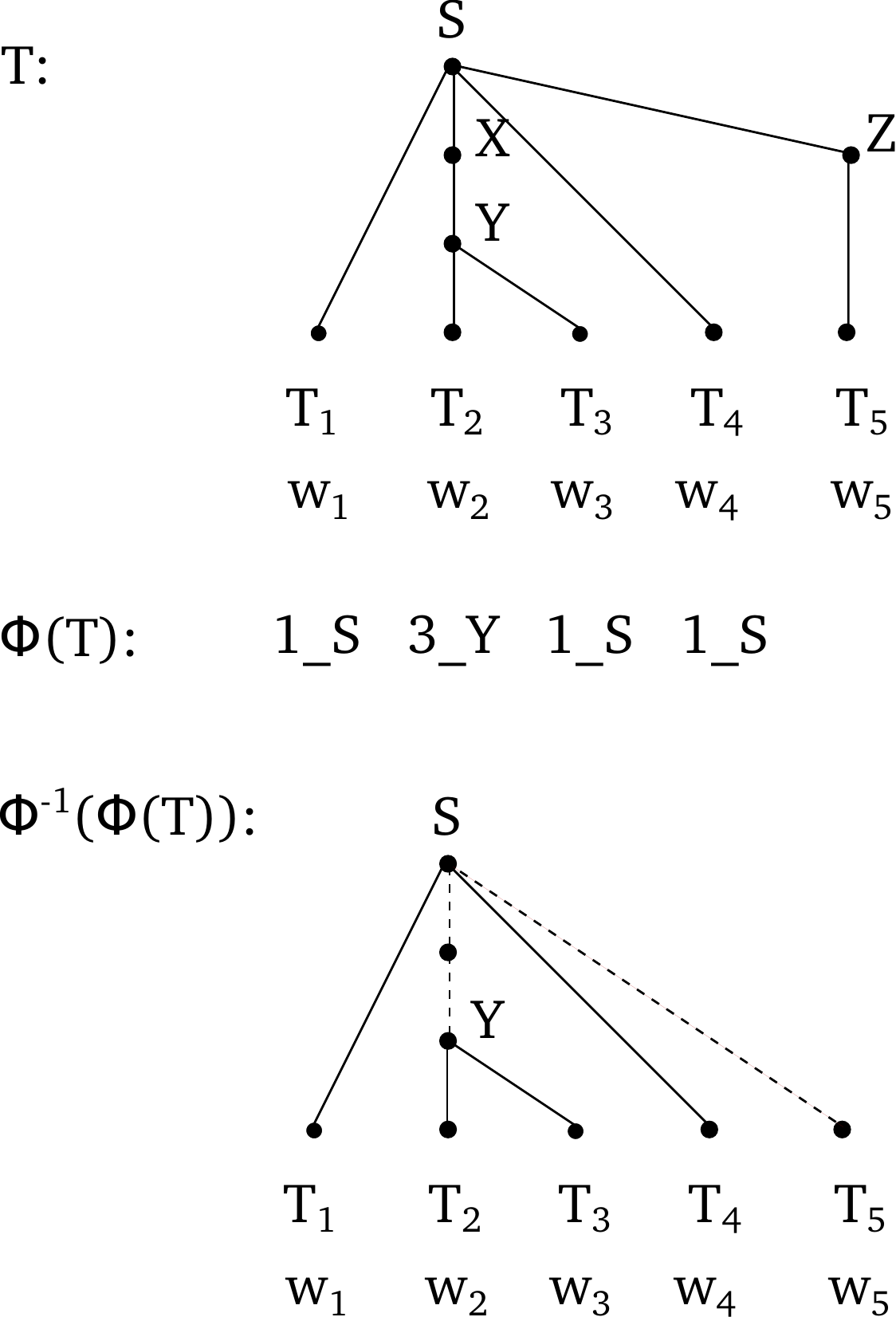}
\caption{\label{f-unary-example} An example of a tree that cannot be directly linearized with our approach. $w_i$ and $T_i$ abstract over words and PoS tags. Dotted lines represent incorrect branches after applying and inverting our encoding naively without any adaptation for unaries. The nonterminal symbol of the second ancestor of $w_2$ (\textsc{x}) cannot be decoded, as no pair of words have \textsc{x} as their lowest common ancestor. A similar situation can be observed for the closest ancestor of $w_5$ (\textsc{z}).}
\end{figure}

\paragraph{Handling of unary branches} The encoding function $\Phi_{|w|}$ cannot directly assign the nonterminal symbols of unary branches, as there is not any pair of words $(w_i,w_{i+1})$ that have those in common. Figure \ref{f-unary-example} illustrates it with an example. 

It is worth remarking that this is not a limitation of our encoding, but of any encoding that would facilitate constituent parsing as sequence labeling, as the number of nonterminal nodes in a tree with unary branches is not bounded by any function of $|w|$. The fact that our encoding works for trees without unary branches owes to the fact that such a tree cannot have more than $|w|-1$ non-leaf nodes, and therefore it is always possible to encode all of them in labels associated with $|w|-1$ leaf nodes.

To overcome this issue, we follow a collapsing approach, as is common in parsers that need special treatment of unary chains \cite{finkel2008efficient,narayan-cohen:2016:P16-1,ShenDistance2018}. For clarity, we use the name \emph{intermediate unary chains} to refer to unary chains that end up into a nonterminal symbol (e.g. $\textsc{x} \rightarrow \textsc{y}$ in Figure \ref{f-unary-example}) and \emph{leaf unary chains} to name those that yield a PoS tag (e.g. $\textsc{z} \rightarrow \textsc{t}_5$). Intermediate unary chains are collapsed into a chained single symbol, which can be encoded by $\Phi_{|w|}$ as any other nonterminal symbol. 
On the other hand, leaf unary chains
are collapsed together with the PoS tag, but these  cannot be encoded and decoded by relying on $\Phi_{|w|}$, as our encoding assumes a fixed sequence of leaf nodes and does not encode them explicitly. 
To overcome this, we propose two methods:

\begin{enumerate}
\item To use an extra function to enrich the PoS tags before applying our main sequence labeling function. 
This function is of the form $\Psi_{|w|}: V^{|w|} \rightarrow U^{|w|}$, where $U$ is the set of labels of the leaf unary chains (without including the PoS tags) plus a dummy label $\varnothing$.
$\Psi_{|w|}$ maps $w_i$ to $\varnothing$ if there is no leaf unary chain at $w_i$, or to the collapsed label otherwise.
\item To extend our encoding function
to predict them as a part of our labels $l_i$, by transforming them into 3-tuples $(n_i,c_i,u_i)$ where $u_i$ encodes the leaf unary chain collapsed label for $w_i$, if there is any, or none otherwise. We call this extended encoding function $\Phi{\,'}_{|w|}$.
\end{enumerate}

The former requires to run two passes of sequence labeling to deal with leaf unary chains. The latter avoids this, but the number of labels is larger and sparser. In \S \ref{section-experiments} we discuss how these two approaches behave in terms of accuracy and speed.

\paragraph{Non-surjectivity} 

Our encoding, as defined formally in Section \ref{sec:encoding}, is injective but not surjective, i.e., not every sequence of $|w|-1$ labels of the form $(n_i,c_i)$ corresponds to a tree in $T_{|w|}$. In particular, there are two situations where a label sequence formally has no tree, and thus $\Phi_{|w|}^{-1}$ is not formally defined and we have to use extra heuristics or processing to define it:
\begin{itemize}
\item Sequences with conflicting nonterminals. A nonterminal can be the lowest common ancestor of more than two pairs of contiguous words when branches are non-binary. For example, in the tree in Figure \ref{f-running-example}, the lowest common ancestor of both ``the'' and ``red'' and of ``red'' and ``toy'' is the same $NP$ node. This translates into ${c_4=\mathit{NP}}$, ${c_5=\mathit{NP}}$ in the label sequence. If we take that sequence and set ${c_5=\mathit{VP}}$, we obtain a label sequence that does not strictly correspond to the encoding of any tree, as it contains a contradiction: two elements referencing the same node indicate different nonterminal labels. In practice, this problem is trivial to solve: when a label sequence encodes several conflicting nonterminals at a given position in the tree, we compute $\Phi_{|w|}^{-1}$ using the first such nonterminal and ignoring the rest.
\item Sequences that produce unary structures. There are sequences of values $n_i$ that do not correspond to a tree in $T_{|w|}$ because the only tree structure satisfying the common ancestor conditions of their values (the one built by generating the string of $\beta_i$s in the injectivity proof) contains unary branchings, causing the problem described above where we do not have a specification for every nonterminal. An example of this is the sequence $(1,S),(3,Y),(1,S),(1,S)$ in absolute scaling, that was introduced in Figure \ref{f-unary-example}. In practice, as unary chains have been previously collapsed, any generated unary node is considered as not valid and removed.
\end{itemize}

\section{Sequence Labeling}\label{section-seq2seq}

Sequence labeling is an structured prediction task that generates an output label for every token in an input sequence \cite{ReiSoeLabeling}. Examples of practical tasks that can be formulated under this framework in natural language processing are PoS tagging, chunking or named-entity recognition, which are in general fast. However, 
to our knowledge, there is no previous work on sequence labeling methods for constituent parsing, as an encoding allowing it was lacking so far.

In this work, we consider a range of methods ranging from traditional models to state-of-the-art neural models for sequence labeling, to test whether they are valid to train constituency-based parsers following our approach. We give the essential details needed to comprehend the core of each approach, but will mainly treat them as black boxes, referring the reader to the references for a careful and detailed mathematical analysis of each method. Appendix \ref{appendix-setup} specifies additional hyper-parameters for the tested models.

\paragraph{Preprocessing} We add to every sentence both beginning and end tokens.

\subsection{Traditional Sequence Labeling Methods}

We consider two baselines to train our prediction function $F_{|w|,\theta}$, based on popular sequence labeling methods used in \textsc{nlp} problems, such as PoS tagging or shallow parsing \cite{Schmid1994Part,sha2003shallow}.

\paragraph{Conditional Random Fields} \cite{lafferty2001conditional} Let \textsc{crf}$_{|w|,\theta}$ be its prediction function, a \textsc{crf} model computes conditional probability distributions of the form $p(\vec{l},\vec{w})$ such that \textsc{crf}$_\theta(\vec{w})$ = $\vec{l}$ = $\argmax_{\vec{l'}} p(\vec{l'},\vec{w})$. In our work, the inputs to the \textsc{crf} are words and PoS tags. To represent a word $w_i$, we are using information of the word itself and also contextual information from $\vec{w}_{[i-1:i+1]}$.\footnote{We tried contextual information beyond the immediate previous and next word, but the performance was similar.} In particular:
\begin{itemize}
\item We extract the word form (lowercased), the PoS tag and its prefix of length 2, from $\vec{w}_{[i-1:i+1]}$. For these words we also include binary features: whether it is the first word, the last word, a number, whether the word is capitalized or uppercased.
\item Additionally, for $w_i$ we look at the suffixes of both length 3 and 2 (i.e. $w_{i[-3:]}$ and $w_{i[-2:]}$).
\end{itemize}

To build our \textsc{CRF} models, we relied on the \texttt{sklearn-crfsuite} library\footnote{https://sklearn-crfsuite.readthedocs.io/en/latest/}.

\paragraph{MultiLayer Perceptron} \cite{rosenblatt1958perceptron} We use one hidden layer. Let \textsc{mlp}$_{|w|,\theta}$ be its prediction function, it treats sequence labeling as a set of independent predictions, one per word. The prediction for a word is computed as $softmax(\vec{W_2} \cdot relu(\vec{W_1} \cdot \vec{x} + \vec{b_1}) + \vec{b_2})$, where $\vec{x}$ is the input vector and $W_i$ and $b_i$ the weights and biases to be learned at layer $i$.
We consider both a discrete (\textsc{mlp}$_d$) and an embedded (\textsc{mlp}$_e$) perceptron. For the former, we use as inputs the same set of features as for the \textsc{crf}. For the latter, the vector $\vec{x}$ for $w_i$ is defined as a concatenation of word and PoS tag embeddings from $\vec{w}_{[i-2:i+2]}$.\footnote{In contrast to the discrete input, larger contextual information was useful.}

To build our \textsc{mlp}s, we relied on \texttt{keras}.\footnote{https://keras.io/}

\subsection{Sequence Labeling Neural Models}

We are using \textsc{ncrf}pp\footnote{https://github.com/jiesutd/NCRFpp, with PyTorch.}, a sequence labeling framework  based on recurrent neural networks (\textsc{rnn}) \cite{yang2017ncrf}, and more specifically on bidirectional short-term memory networks \cite{hochreiter1997long}, which have been successfully applied to problems such as PoS tagging or dependency parsing \cite{plank2016multilingual,kiperwasser2016simple}. Let \textsc{lstm}$(\vec{x})$ be an abstraction of a standard long short-term memory network that processes the sequence $\vec{x}=[\vec{x}_1,...,\vec{x}_{|\vec{x}|}]$, then a \textsc{bilstm} encoding of its $i$th element, \textsc{bilstm}$(\vec{x},i)$ is  defined as:

\begin{center}

\textsc{bilstm}$(\vec{x},i)$ = $\vec{h}_i$ = $\vec{h}^l_i \circ \vec{h}^r_i $ = $\textsc{lstm$^l$}(\vec{x}_{[1:i]}) \circ \textsc{lstm$^r$}(\vec{x}_{[|\vec{x}|:i]})$
\end{center}

In the case of multilayer \textsc{bilstm's}, the time-step outputs of the \textsc{bilstm}$_m$ are fed as input to  the \textsc{bilstm}$_{m+1}$. The output label for each $w_i$ is finally predicted as $softmax(W \cdot \vec{h}_i + b)$.

\begin{figure}[hbtp]
\centering
\vspace{-0.7cm}
\includegraphics[width=0.9\columnwidth]{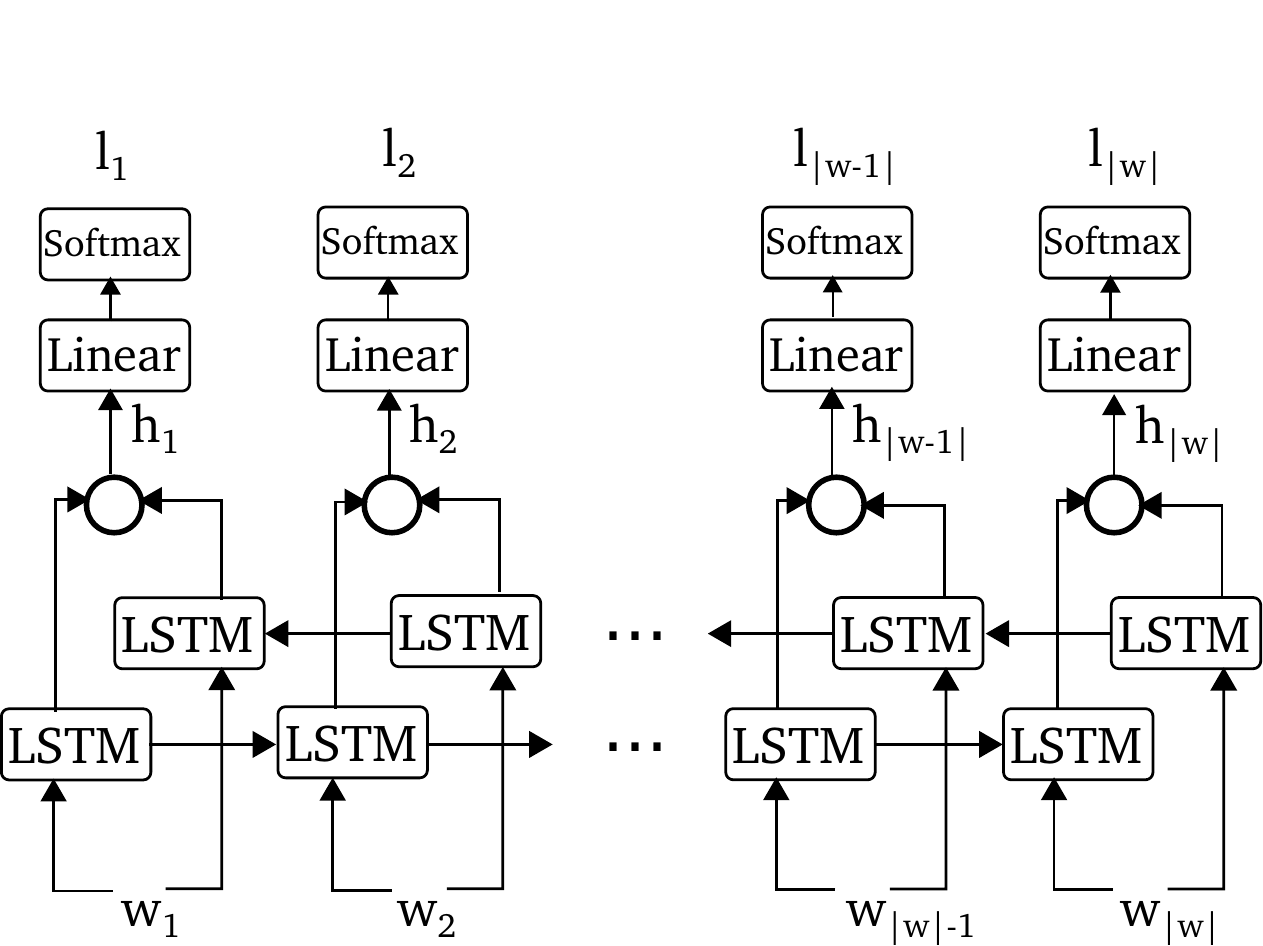}
\caption{\label{f-ncrf} Architecture of the neural model}
\end{figure}

Given a sentence $[w_1,w_2,...,w_{|w|}]$, the input to the sequence model is a sequence of embeddings $[\vec{w_1},\vec{w_2},...,\vec{w_{|w|}}]$ where each $\vec{w_i}= \vec{w}_i\, \circ\, \vec{p}_{i}\, \circ\, \vec{ch}_i$, such that $\vec{w}_i$ and $\vec{p}_i$ are a word and a PoS tag embedding, and $\vec{ch}_i$ is a word embedding obtained from an initial character embedding layer, also based on a \textsc{bilstm}. Figure \ref{f-ncrf} shows the architecture of the network.

\section{Experiments}\label{section-experiments}

We report results on models trained using the relative scale encoding and the special tag (\textsc{root},$c_i$). As a reminder, to deal also with leaf unary chains, we proposed two methods in \S \ref{section-limitations}: to predict them relying both on the encoding functions $\Phi_{|w|}$ and $\Psi_{|w|}$, or to predict them as a part of an enriched label predicted by the function $\Phi{\,'}_{|w|}$. For clarity, we are naming these models with the superscripts $^{\Psi,\Phi}$ and $^{\Phi\,'}$, respectively. 

\paragraph{Datasets} We use the Penn Treebank \cite{Marcus:1994} and its official splits: Sections 2 to 21 for training, 22 for development and 23 for testing. For the Chinese Penn Treebank \cite{xue2005penn}: articles 001- 270 and 440-1151 are used
for training, articles 301-325 for development, and articles 271-300 for testing. We use the version of the corpus with the predicted PoS tags of \newcite{DyerRecurrent2016}. We train the $\Phi$ models based on the predicted output by the corresponding $\Psi$ model.

\paragraph{Metrics} We use the F-score from the \textsc{evalb} script using COLLINS.prm as the parameter file. Speed is measured in sentences per second. 
We briefly comment on the accuracy (percentage of correctly predicted labels, no symbol excluded here) of our baselines.

\paragraph{Source code} It can be found at \url{https://github.com/aghie/tree2labels}

\paragraph{Hardware} The models are run on a single thread of a CPU\footnote{An Intel(R) Core(TM) i7-7700 CPU @ 3.60GHz.} and on a consumer-grade GPU\footnote{A GeForce GTX 1080.}.
In sequence-to-sequence work \cite{vinyals2015grammar} the authors 
use
a multi-core CPU (the number of threads was not specified), 
while we provide results on a single core for easier comparability. 
Parsing sentences on a CPU can be framed as an ``embarrassingly parallel'' problem \cite{hall2014sparser}, so speed can be made to scale linearly with the number of cores. We use the same batch size as \citet{vinyals2015grammar} for testing (128).\footnote{A larger batch will likely result in faster parsing when executing the model on a \textsc{gpu}, but not necessarily on a \textsc{cpu}.}

\subsection{Results}

Table \ref{table-baselines} shows the performance of our baselines on the \textsc{ptb} development set. It is worth noting that since we are using different libraries to train the models, these might show some differences in terms of performance/speed beyond those expected in theory.
For the \textsc{bilstm} model we test:
\begin{itemize}
\item \textsc{bilstm}$_{m=1}$: It does not use pretrained word embeddings nor character embeddings. The number of layers $m$ is set to 1.
\item \textsc{bilstm}$_{m=1,e}$: It adds pretrained word embeddings from GloVe \cite{pennington2014glove} for English and from the Gigaword corpus for Chinese \cite{Liu2017InOrder}.
\item \textsc{bilstm}$_{m=1,e,ch}$: It includes character embeddings processed through a \textsc{bilstm}.
\item \textsc{bilstm}$_{m=2,e}$: $m$ is set to 2. No character embeddings. 

\item \textsc{bilstm}$_{m=2,e,ch}$: $m$ is set to 2.
\end{itemize}

\begin{table}[hbtp]
\begin{center}
\small
\begin{tabular}{|lrrrr|}
\hline
  \multirow{2}{*}{\bf Model} &  \multirow{2}{*}{\bf F-score} &  \multirow{2}{*}{\bf Acc.} & \bf Sent/s &\bf Sent/s \\
 &&&\bf (\textsc{cpu})&\bf (\textsc{gpu})\\
 \hline
  \textsc{crf}$^{\Psi,\Phi}$ &62.5&63.9&83&-\\
  \textsc{mlp}$^{\Psi,\Phi}_d$ &74.8&78.0&16&49\\
  \textsc{mlp}$^{\Psi,\Phi}_e$ &76.6&79.3&503&666\\

 \textsc{crf}$^{\Phi\,'}$ &62.3&65.4&6&-\\
 \textsc{mlp}$^{\Phi\,'}_d$ &74.2&78.0&31&95\\
 \textsc{mlp}$^{\Phi\,'}_e$ &77.2&79.7&342&890\\
 \hline
 \textsc{bilstm}$^{\Psi,\Phi}_{m=1}$&88.1&88.9&144&541\\
 \textsc{bilstm}$^{\Psi,\Phi}_{m=1,e}$&89.2&89.8&144&543\\
 \textsc{bilstm}$^{\Psi,\Phi}_{m=1,e,ch}$&89.5&90.0&120&456\\
 \textsc{bilstm}$^{\Psi,\Phi}_{m=2,e}$&90.7&90.7&72&476\\
 \textsc{bilstm}$^{\Psi,\Phi}_{m=2,e,ch}$&90.7&90.9&65&405\\
 \hline
  \textsc{bilstm}$^{\Phi\,'}_{m=1}$&88.3&89.3&206&941\\
 \textsc{bilstm}$^{\Phi\,'}_{m=1,e}$&89.5&90.1&209&957\\
 \textsc{bilstm}$^{\Phi\,'}_{m=1,e,ch}$&89.0&90.0&180&808\\
 \textsc{bilstm}$^{\Phi\,'}_{m=2,e}$&90.8&90.9&119&842\\
 \textsc{bilstm}$^{\Phi\,'}_{m=2,e,ch}$&90.6&90.9&109&716\\
 \hline

\end{tabular}
\end{center}
\caption{\label{table-baselines} Performance of the proposed sequence labeling methods on the development set of the \textsc{ptb}. For the \textsc{crf} models the complexity is quadratic with respect to the number of labels, which causes \textsc{crf}$^{\Phi\,'}$ to be particularly slow.
}
\end{table}




\begin{table*}[h!]
\begin{center}
\small
\begin{tabular}{|lcccccc|}
\hline

 \bf \multirow{2}{*}{Model} & \bf \multirow{2}{*}{Testbed} & \multicolumn{2}{c}{\bf CPU Run} & \multicolumn{2}{c}{\bf GPU Run}  & \bf F-score\\
  &  &\bf \#Cores &\bf Sents/s &\bf \#GPU & \bf Sents/s & \\
\hline
 \multicolumn{6}{l}{\textcolor{gray}{Sequence labeling}}\\

 \hline
 \textsc{mlp}$^{\Psi,\Phi}_e$&WSJ23&1&501&1&669&75.8\\
  \textsc{mlp}$^{\Phi\,'}_e$&WSJ23&1&349&1&929&76.7\\
  \textsc{bilstm}$^{\Psi,\Phi}_{m=1,e}$&WSJ23&1&148&1&581&88.9\\
 \textsc{bilstm}$^{\Phi\,'}_{m=1,e}$&WSJ23&1&221&1&1016&89.1\\ 
 \textsc{bilstm}$^{\Psi,\Phi}_{m=2,e,ch}$&WSJ23&1&66&1&434&90.6\\
 \textsc{bilstm}$^{\Phi\,'}_{m=2,e,ch}$&WSJ23&1&115&1&780&90.7\\
 \textsc{bilstm}$^{\Psi,\Phi}_{m=2,e}$&WSJ23&1&74&1&506&90.7\\
 \textsc{bilstm}$^{\Phi\,'}_{m=2,e}$&WSJ23&1&126&1&898&90.7\\
 \hline

 \multicolumn{6}{l}{\textcolor{gray}{Sequence-to-sequence}}\\
 \hline 3-layer \textsc{lstm} &WSJ 23&&&&&$<$70\\
\multirow{3}{*}{3-layer \textsc{lstm} + Attention$^\diamond$} &\multirow{3}{*}{WSJ 23}&Multi-core&\multirow{3}{*}{120}&&&\multirow{3}{*}{88.3}\\
&&(number not&&&&\\
\cite{vinyals2015grammar}&&specified)&&&&\\
\hline
\multicolumn{6}{l}{\textcolor{gray}{Constituency parsing as dependency parsing}}\\
\hline
\newcite{Fer2015Parsing}$^\diamond$&WSJ23&1&41&&&90.2\\
\hline
\multicolumn{6}{l}{\textcolor{gray}{Chart-based parsers}}\\
\hline
\newcite{charniak2000maximum}$^\ast$&WSJ23&1&6&&&89.5\\
\newcite{petrov2007improved}$^\ast$&WSJ23&1&6&&&90.1\\
\newcite{stern2017minimal}$^\diamond$&WSJ23&16\textsuperscript{*}&20&&&91.8\\
\newcite{Kitaev2018Constituency}&\multirow{2}{*}{WSJ23}&&&\multirow{2}{*}{2}&\multirow{2}{*}{70}&\multirow{2}{*}{95.1}\\
+ELMo \cite{peters2018deep}$^\diamond$&&&&&&\\
\hline
\multicolumn{6}{l}{\textcolor{gray}{Chart-based parsers with GPU-specific implementation}}\\
\hline
\newcite{canny2013multi}$^\diamond$&WSJ($<$30)&&&1&250&\\
\newcite{hall2014sparser}$^\diamond$&WSJ($<$40)&&&1&404&\\

\hline
\multicolumn{6}{l}{\textcolor{gray}{Transition-based and other greedy constituent parsers}}\\
\hline
\newcite{zhu2013fast}$^\diamond$&WSJ23&1&101&&&89.9\\
\newcite{zhu2013fast}+Padding$^\diamond$&WSJ23&1&90&&&90.4\\
\newcite{DyerRecurrent2016}$^\rhd$&WSJ23&1&17&&&91.2\\
\defcitealias{Fer2018Faster}{Fern\'andez and G\'omez-Rodr\'iguez (2018)}\citetalias{Fer2018Faster}$^\diamond$&WSJ23&1&18&&&91.7\\
\newcite{stern2017minimal}$^\diamond$&WSJ23&16\textsuperscript{*}&76&&&91.8\\
\newcite{Liu2017InOrder}&WSJ23&&&&&91.8\\
\newcite{ShenDistance2018}&WSJ23&&&1&111&91.8\\
\hline

\end{tabular}
\end{center}
\caption{\label{table-sota} Comparison against the state of the art.\textsuperscript{*}\newcite{stern2017minimal} report that they use a 16-core machine, but sentences are processed one-at-a-time. Hence, they do not exploit inter-sentence parallelism, but they may gain some speed from intra-sentence parallelism. $\diamond$ indicates the that the speed was reported in the paper itself. $\ast$ and $\rhd$ indicate that the speeds were extracted from \newcite{zhu2013fast} and \defcitealias{Fer2018Faster}{Fern\'andez and G\'omez-Rodr\'iguez (2018)}\citetalias{Fer2018Faster}.
}
\end{table*}

\begin{table}[bpth]
\begin{center}
\small
\begin{tabular}{|lc|}
\hline
 \bf Model & \bf F-score   \\
 \hline
 
 \textsc{mlp}$^{\Psi,\Phi}_e$ &63.1\\
 \textsc{mlp}$^{\Phi\,'}_e$ &64.4\\
 \textsc{bilstm}$^{\Psi,\Phi}_{m=2,e,ch}$&84.4\\
 \textsc{bilstm}$^{\Phi\,'}_{m=2,e,ch}$&84.1\\
 \textsc{bilstm}$^{\Psi,\Phi}_{m=2,e}$&84.4\\
 \textsc{bilstm}$^{\Phi\,'}_{m=2,e}$&83.1\\
 
 \hline
 \hline
 \newcite{zhu2013fast}&82.6\\
 \newcite{zhu2013fast}+P&83.2\\
 \newcite{DyerRecurrent2016}&84.6\\
 \newcite{Liu2017InOrder}&86.1\\
 \newcite{ShenDistance2018}&86.5\\
  \defcitealias{Fer2018Faster}{Fern\'andez and G\'omez-Rodr\'iguez (2018)}\citetalias{Fer2018Faster}&86.8\\
\hline

\end{tabular}
\end{center}
\caption{\label{table-sota-chinese} Performance on the \textsc{ctb} test set}
\end{table}

The ${\Psi,\Phi}$ and the ${\Phi\,'}$ models obtain similar F-scores.
When it comes to speed, the \textsc{bilstm}s$^{\Phi\,'}$ are notably faster than the \textsc{bilstm}s$^{\Psi,\Phi}$.
$\Phi\,'$ models are expected to be more efficient, as leaf unary chains are handled implicitly. 
In practice, $\Phi\,'$ is a more expensive function to compute than the original $\Phi$, since the number of output labels is significantly larger, which reduces the expected gains with respect to the $\Psi,\Phi$ models. It is worth noting that our encoding is useful to train an \textsc{mlp}$_e$ with a decent sense of phrase structure, while being very fast. Paying attention to the differences between F-score and Accuracy for each baseline, we notice the gap between them is larger for \textsc{crf}s and \textsc{mlp}s. This shows the difficulties that these methods have, in comparison to the \textsc{bilstm} approaches, to predict the correct label when a word $w_{i+1}$ has few common ancestors with $w_i$. For example,
let \textsc{-10x} be the right (relative scale) label between $w_i$ and $w_{i+1}$, and let $l_1$=\textsc{-1x} and $l_2$=\textsc{-9x} be two possible wrong labels. In terms of accuracy
it is the same that a model predicts $l1$ or $l2$, but in terms of constituent F-score, the first will be much worse, as many closed parentheses will remain unmatched. 

Tables \ref{table-sota} and \ref{table-sota-chinese} compare our best models against the state of the art on the \textsc{ptb} and \textsc{ctb} test sets. The performance corresponds to models without reranking strategies, unless otherwise specified.

\section{Discussion} We are not aware of work that reduces constituency parsing to sequence labeling. The work that can be considered as the closest to ours is that of \newcite{vinyals2015grammar}, who address it as a sequence-to-sequence problem, where the output sequence has variable/unknown length. In this context, even a one hidden layer perceptron outperforms their 3-layer \textsc{lstm} model without attention, while parsing hundreds of sentences per second. Our best models also outperformed their 3-layer \textsc{lstm} model with attention and even a simple \textsc{bilstm} model with pre-trained GloVe embeddings obtains a similar performance.
In terms of F-score, the proposed sequence labeling baselines still lag behind mature shift-reduce and chart parsers. In terms of speed, they are clearly faster than both CPU and GPU chart parsers and are at least on par with the fastest shift-reduce ones. Although with significant loss of accuracy, if phrase-representation is needed in large-scale tasks where the speed of current systems makes parsing infeasible \cite{gomez2017towards,gomez2017important}, 
we can use the simpler, less accurate models to
get speeds well above any parser reported to date.

It is also worth noting that in their recent work, published while this manuscript was under review, \newcite{ShenDistance2018} developed a mapping of binary trees with $n$ leaves to sequences of $n-1$ integers \cite[Algorithm~1]{ShenDistance2018}. This encoding is different from the ones presented here, as it is based on the height of lowest common ancestors in the tree, rather than their depth. While their purpose is also different from ours, as they use this mapping to generate training data for a parsing algorithm based on recursive partitioning using real-valued distances, their encoding could also be applied with our sequence labeling approach. However, it has the drawback that it only supports binarized trees, and some of its theoretical properties are worse for our goal, as the way to define the inverse of an arbitrary label sequence can be highly ambiguous: for example, a sequence of $n-1$ equal labels in this encoding can represent \emph{any} binary tree with $n$ leaves.

\section{Conclusion}

We presented a new parsing paradigm, based on a reduction of constituency parsing to sequence labeling. We first described a linearization function to transform a constituent tree (with $n$ leaves) into a sequence of $n-1$ labels that encodes it. We proved that this encoding function is total and injective for any tree without unary branches. We also discussed its limitations: how to deal with unary branches and non-surjectivity, and showed how these can be solved. We finally proposed a set of fast and strong baselines.

\section*{Acknowledgments}

This work has received funding from the European
Research Council (ERC), under the European
Union's Horizon 2020 research and innovation
programme (FASTPARSE, grant agreement No
714150), from the TELEPARES-UDC project
(FFI2014-51978-C2-2-R) and the ANSWER-ASAP project (TIN2017-85160-C2-1-R) from MINECO, and from Xunta de Galicia (ED431B 2017/01). 
We gratefully acknowledge NVIDIA Corporation for the donation of a GTX Titan X GPU.

\bibliography{emnlp2018}

\begin{thebibliography}{38}
\expandafter\ifx\csname natexlab\endcsname\relax\def\natexlab#1{#1}\fi

\bibitem[{Canny et~al.(2013)Canny, Hall, and Klein}]{canny2013multi}
John Canny, David Hall, and Dan Klein. 2013.
\newblock A multi-teraflop constituency parser using {GPU}s.
\newblock In \emph{Proceedings of the 2013 Conference on Empirical Methods in
  Natural Language Processing}, pages 1898--1907.

\bibitem[{Charniak(2000)}]{charniak2000maximum}
Eugene Charniak. 2000.
\newblock A maximum-entropy-inspired parser.
\newblock In \emph{Proceedings of the 1st North American chapter of the
  Association for Computational Linguistics conference}, pages 132--139.
  Association for Computational Linguistics.

\bibitem[{Choe and Charniak(2016)}]{ChoeChar2016}
Do~Kook Choe and Eugene Charniak. 2016.
\newblock Parsing as language modeling.
\newblock In \emph{Proceedings of the 2016 Conference on Empirical Methods in
  Natural Language Processing}, pages 2331--2336, Austin, Texas. Association
  for Computational Linguistics.

\bibitem[{Collins(1997)}]{collins1997three}
Michael Collins. 1997.
\newblock Three generative, lexicalised models for statistical parsing.
\newblock In \emph{Proceedings of the eighth conference on European chapter of
  the Association for Computational Linguistics}, pages 16--23. Association for
  Computational Linguistics.

\bibitem[{Dyer et~al.(2016)Dyer, Kuncoro, Ballesteros, and
  Smith}]{DyerRecurrent2016}
Chris Dyer, Adhiguna Kuncoro, Miguel Ballesteros, and Noah~A. Smith. 2016.
\newblock Recurrent neural network grammars.
\newblock In \emph{Proceedings of the 2016 Conference of the North American
  Chapter of the Association for Computational Linguistics: Human Language
  Technologies}, pages 199--209. Association for Computational Linguistics.

\bibitem[{{Fern{\'a}ndez-Gonz{\'a}lez} and
  {G{\'o}mez}-Rodr{\'i}guez(2018)}]{Fer2018Faster}
Daniel {Fern{\'a}ndez-Gonz{\'a}lez} and Carlos {G{\'o}mez}-Rodr{\'i}guez. 2018.
\newblock {Faster Shift-Reduce Constituent Parsing with a Non-Binary, Bottom-Up
  Strategy}.
\newblock \emph{ArXiv e-prints}.

\bibitem[{Fern{\'a}ndez-Gonz{\'a}lez and Martins(2015)}]{Fer2015Parsing}
Daniel Fern{\'a}ndez-Gonz{\'a}lez and Andr{\'e} F.~T. Martins. 2015.
\newblock Parsing as reduction.
\newblock In \emph{Proceedings of the 53rd Annual Meeting of the Association
  for Computational Linguistics and the 7th International Joint Conference on
  Natural Language Processing (Volume 1: Long Papers)}, pages 1523--1533.
  Association for Computational Linguistics.

\bibitem[{Finkel et~al.(2008)Finkel, Kleeman, and
  Manning}]{finkel2008efficient}
Jenny~Rose Finkel, Alex Kleeman, and Christopher~D Manning. 2008.
\newblock Efficient, feature-based, conditional random field parsing.
\newblock \emph{Proceedings of ACL-08: HLT}, pages 959--967.

\bibitem[{G{\'o}mez-Rodr{\'\i}guez(2017)}]{gomez2017towards}
Carlos G{\'o}mez-Rodr{\'\i}guez. 2017.
\newblock Towards fast natural language parsing: {FASTPARSE ERC} {S}tarting
  {G}rant.
\newblock \emph{Procesamiento del Lenguaje Natural}, 59.

\bibitem[{G{\'o}mez-Rodr{\'i}guez et~al.(2017)G{\'o}mez-Rodr{\'i}guez,
  Alonso-Alonso, and Vilares}]{gomez2017important}
Carlos G{\'o}mez-Rodr{\'i}guez, Iago Alonso-Alonso, and David Vilares. 2017.
\newblock How important is syntactic parsing accuracy? {A}n empirical
  evaluation on rule-based sentiment analysis.
\newblock \emph{Artificial Intelligence Review}.

\bibitem[{Hall et~al.(2014)Hall, Berg-Kirkpatrick, and Klein}]{hall2014sparser}
David Hall, Taylor Berg-Kirkpatrick, and Dan Klein. 2014.
\newblock Sparser, better, faster {GPU} parsing.
\newblock In \emph{Proceedings of the 52nd Annual Meeting of the Association
  for Computational Linguistics (Volume 1: Long Papers)}, volume~1, pages
  208--217.

\bibitem[{Hochreiter and Schmidhuber(1997)}]{hochreiter1997long}
Sepp Hochreiter and J{\"u}rgen Schmidhuber. 1997.
\newblock Long short-term memory.
\newblock \emph{Neural computation}, 9(8):1735--1780.

\bibitem[{Kiperwasser and Goldberg(2016)}]{kiperwasser2016simple}
Eliyahu Kiperwasser and Yoav Goldberg. 2016.
\newblock Simple and accurate dependency parsing using bidirectional {LSTM}
  feature representations.
\newblock \emph{Transactions of the Association for Computational Linguistics},
  4:313--327.

\bibitem[{Kitaev and Klein(2018)}]{Kitaev2018Constituency}
Nikita Kitaev and Dan Klein. 2018.
\newblock Constituency parsing with a self-attentive encoder.
\newblock In \emph{Proceedings of the 56th Annual Meeting of the Association
  for Computational Linguistics (Volume 1: Long Papers)}, Melbourne, Australia.
  Association for Computational Linguistics.

\bibitem[{Kummerfeld et~al.(2012)Kummerfeld, Hall, Curran, and
  Klein}]{kummerfeldCorral}
Jonathan~K. Kummerfeld, David Hall, James~R. Curran, and Dan Klein. 2012.
\newblock Parser showdown at the {W}all {S}treet corral: An empirical
  investigation of error types in parser output.
\newblock In \emph{Proceedings of the 2012 Joint Conference on Empirical
  Methods in Natural Language Processing and Computational Natural Language
  Learning}, pages 1048--1059, Jeju Island, Korea. Association for
  Computational Linguistics.

\bibitem[{Lafferty et~al.(2001)Lafferty, McCallum, and
  Pereira}]{lafferty2001conditional}
John~D. Lafferty, Andrew McCallum, and Fernando C.~N. Pereira. 2001.
\newblock Conditional random fields: Probabilistic models for segmenting and
  labeling sequence data.
\newblock In \emph{Proceedings of the Eighteenth International Conference on
  Machine Learning}, ICML '01, pages 282--289, San Francisco, CA, USA. Morgan
  Kaufmann Publishers Inc.

\bibitem[{{Liu} and {Zhang}(2017)}]{Liu2017InOrder}
J.~{Liu} and Y.~{Zhang}. 2017.
\newblock {In-Order Transition-based Constituent Parsing}.
\newblock \emph{ArXiv e-prints}.

\bibitem[{Marcus et~al.(1994)Marcus, Kim, Marcinkiewicz, MacIntyre, Bies,
  Ferguson, Katz, and Schasberger}]{Marcus:1994}
Mitchell Marcus, Grace Kim, Mary~Ann Marcinkiewicz, Robert MacIntyre, Ann Bies,
  Mark Ferguson, Karen Katz, and Britta Schasberger. 1994.
\newblock The {P}enn {T}reebank: Annotating predicate argument structure.
\newblock In \emph{Proceedings of the Workshop on Human Language Technology},
  HLT '94, pages 114--119, Stroudsburg, PA, USA. Association for Computational
  Linguistics.

\bibitem[{Narayan and Cohen(2016)}]{narayan-cohen:2016:P16-1}
Shashi Narayan and Shay~B. Cohen. 2016.
\newblock Optimizing spectral learning for parsing.
\newblock In \emph{Proceedings of the 54th Annual Meeting of the Association
  for Computational Linguistics (Volume 1: Long Papers)}, pages 1546--1556,
  Berlin, Germany. Association for Computational Linguistics.

\bibitem[{Nivre(2003)}]{nivre2003efficient}
Joakim Nivre. 2003.
\newblock An efficient algorithm for projective dependency parsing.
\newblock In \emph{Proceedings of the 8th International Workshop on Parsing
  Technologies (IWPT)}, pages 149--160.

\bibitem[{Pennington et~al.(2014)Pennington, Socher, and
  Manning}]{pennington2014glove}
Jeffrey Pennington, Richard Socher, and Christopher Manning. 2014.
\newblock Glove: Global vectors for word representation.
\newblock In \emph{Proceedings of the 2014 Conference on Empirical Methods in
  Natural Language Processing (EMNLP)}, pages 1532--1543.

\bibitem[{Peters et~al.(2018)Peters, Neumann, Iyyer, Gardner, Clark, Lee, and
  Zettlemoyer}]{peters2018deep}
Matthew Peters, Mark Neumann, Mohit Iyyer, Matt Gardner, Christopher Clark,
  Kenton Lee, and Luke Zettlemoyer. 2018.
\newblock Deep contextualized word representations.
\newblock In \emph{Proceedings of the 2018 Conference of the North American
  Chapter of the Association for Computational Linguistics: Human Language
  Technologies, Volume 1 (Long Papers)}, pages 2227--2237. Association for
  Computational Linguistics.

\bibitem[{Petrov et~al.(2006)Petrov, Barrett, Thibaux, and
  Klein}]{petrov2006learning}
Slav Petrov, Leon Barrett, Romain Thibaux, and Dan Klein. 2006.
\newblock Learning accurate, compact, and interpretable tree annotation.
\newblock In \emph{Proceedings of the 21st International Conference on
  Computational Linguistics and the 44th annual meeting of the Association for
  Computational Linguistics}, pages 433--440. Association for Computational
  Linguistics.

\bibitem[{Petrov and Klein(2007)}]{petrov2007improved}
Slav Petrov and Dan Klein. 2007.
\newblock Improved inference for unlexicalized parsing.
\newblock In \emph{Human Language Technologies 2007: The Conference of the
  North American Chapter of the Association for Computational Linguistics;
  Proceedings of the Main Conference}, pages 404--411.

\bibitem[{Plank et~al.(2016)Plank, S{\o}gaard, and
  Goldberg}]{plank2016multilingual}
Barbara Plank, Anders S{\o}gaard, and Yoav Goldberg. 2016.
\newblock Multilingual part-of-speech tagging with bidirectional long
  short-term memory models and auxiliary loss.
\newblock In \emph{Proceedings of the 54th Annual Meeting of the Association
  for Computational Linguistics (Volume 2: Short Papers)}, pages 412--418,
  Berlin, Germany. Association for Computational Linguistics.

\bibitem[{Rei and S{\o}gaard(2018)}]{ReiSoeLabeling}
Marek Rei and Anders S{\o}gaard. 2018.
\newblock Zero-shot sequence labeling: Transferring knowledge from sentences to
  tokens.
\newblock In \emph{Proceedings of the 2018 Conference of the North American
  Chapter of the Association for Computational Linguistics: Human Language
  Technologies, Volume 1 (Long Papers)}, pages 293--302. Association for
  Computational Linguistics.

\bibitem[{Rosenblatt(1958)}]{rosenblatt1958perceptron}
Frank Rosenblatt. 1958.
\newblock The perceptron: a probabilistic model for information storage and
  organization in the brain.
\newblock \emph{Psychological review}, 65(6):386.

\bibitem[{Sagae and Lavie(2005)}]{sagae2005classifier}
Kenji Sagae and Alon Lavie. 2005.
\newblock A classifier-based parser with linear run-time complexity.
\newblock In \emph{Proceedings of the Ninth International Workshop on Parsing
  Technology}, pages 125--132. Association for Computational Linguistics.

\bibitem[{Schmid(1994)}]{Schmid1994Part}
Helmut Schmid. 1994.
\newblock Part-of-speech tagging with neural networks.
\newblock In \emph{Proceedings of the 15th Conference on Computational
  Linguistics - Volume 1}, COLING '94, pages 172--176, Stroudsburg, PA, USA.
  Association for Computational Linguistics.

\bibitem[{Sha and Pereira(2003)}]{sha2003shallow}
Fei Sha and Fernando Pereira. 2003.
\newblock Shallow parsing with conditional random fields.
\newblock In \emph{Proceedings of the 2003 Conference of the North American
  Chapter of the Association for Computational Linguistics on Human Language
  Technology-Volume 1}, pages 134--141. Association for Computational
  Linguistics.

\bibitem[{Shen et~al.(2018)Shen, Lin, Jacob, Sordoni, Courville, and
  Bengio}]{ShenDistance2018}
Yikang Shen, Zhouhan Lin, Athul~Paul Jacob, Alessandro Sordoni, Aaron
  Courville, and Yoshua Bengio. 2018.
\newblock Straight to the tree: Constituency parsing with neural syntactic
  distance.
\newblock In \emph{Proceedings of the 56th Annual Meeting of the Association
  for Computational Linguistics (Volume 1: Long Papers)}, pages 1171--1180.
  Association for Computational Linguistics.

\bibitem[{Stern et~al.(2017)Stern, Andreas, and Klein}]{stern2017minimal}
Mitchell Stern, Jacob Andreas, and Dan Klein. 2017.
\newblock A minimal span-based neural constituency parser.
\newblock In \emph{Proceedings of the 55th Annual Meeting of the Association
  for Computational Linguistics (Volume 1: Long Papers)}, pages 818--827,
  Vancouver, Canada. Association for Computational Linguistics.

\bibitem[{Vinyals et~al.(2015)Vinyals, Kaiser, Koo, Petrov, Sutskever, and
  Hinton}]{vinyals2015grammar}
Oriol Vinyals, {\L}ukasz Kaiser, Terry Koo, Slav Petrov, Ilya Sutskever, and
  Geoffrey Hinton. 2015.
\newblock Grammar as a foreign language.
\newblock In \emph{Advances in Neural Information Processing Systems}, pages
  2773--2781.

\bibitem[{Wang et~al.(2006)Wang, Sagae, and Mitamura}]{Wang2006}
Mengqiu Wang, Kenji Sagae, and Teruko Mitamura. 2006.
\newblock A fast, accurate deterministic parser for chinese.
\newblock In \emph{Proceedings of the 21st International Conference on
  Computational Linguistics and the 44th Annual Meeting of the Association for
  Computational Linguistics}, ACL-44, pages 425--432, Stroudsburg, PA, USA.
  Association for Computational Linguistics.

\bibitem[{Xue et~al.(2005)Xue, Xia, Chiou, and Palmer}]{xue2005penn}
Naiwen Xue, Fei Xia, Fu-Dong Chiou, and Marta Palmer. 2005.
\newblock The {P}enn {C}hinese {T}reebank: Phrase structure annotation of a
  large corpus.
\newblock \emph{Natural language engineering}, 11(2):207--238.

\bibitem[{Yang and Zhang(2018)}]{yang2017ncrf}
Jie Yang and Yue Zhang. 2018.
\newblock {NCRF}++: An open-source neural sequence labeling toolkit.
\newblock In \emph{Proceedings of ACL 2018, System Demonstrations}, pages
  74--79, Melbourne, Australia. Association for Computational Linguistics.

\bibitem[{Zaremba et~al.(2014)Zaremba, Sutskever, and
  Vinyals}]{zaremba2014recurrent}
Wojciech Zaremba, Ilya Sutskever, and Oriol Vinyals. 2014.
\newblock Recurrent neural network regularization.
\newblock \emph{arXiv preprint arXiv:1409.2329}.

\bibitem[{Zhu et~al.(2013)Zhu, Zhang, Chen, Zhang, and Zhu}]{zhu2013fast}
Muhua Zhu, Yue Zhang, Wenliang Chen, Min Zhang, and Jingbo Zhu. 2013.
\newblock Fast and accurate shift-reduce constituent parsing.
\newblock In \emph{Proceedings of the 51st Annual Meeting of the Association
  for Computational Linguistics (Volume 1: Long Papers)}, volume~1, pages
  434--443.

\end{thebibliography}
\bibliographystyle{acl_natbib_nourl}

\clearpage
\appendix

\section{Setup configuration used to train our sequence labeling methods}\label{appendix-setup}

\paragraph{Conditional Random Fields} We use the default configuration provided together with the \texttt{sklearn-crfsuite} library.

\paragraph{MultiLayer Perceptron} Both the discrete and distributed perceptrons are implemented in \texttt{keras}.
\begin{itemize}
\item \emph{Training hyperparameters} The model is trained up to 30 epochs, with early stopping (patience=4). We use Stochastic Gradient Descent (\textsc{sgd}) to optimize the objective function. The initial learning rate is set to 0.1.
\item \emph{Layer and embedding sizes}. The dimension of the hidden layer is set to 100. For the perceptron fed with embeddings, we use 100 and 20 dimensions to represent a word and its PoS tag, respectively.
\end{itemize}

\paragraph{Bidirectional Long Short-Term Memory} We relied on the \texttt{NCRFpp} framework \cite{yang2017ncrf}.
\begin{itemize}
\item \emph{Training hyperparameters} We use mini-batching (the batch size during training is set to 8). As optimizer, we use \textsc{sgd}, setting the initial learning rate to 0.2, momentum to 0.9 and a linear decay of 0.05. We train the model up to 100 epochs and keep the best performing model in the development set.
\item \emph{Layer and embedding sizes}: We use 100, 30 and 20 dimensions to represent a word, a postag and a character embedding. The output hidden layer from the character embeddings layer is set to 50. The left-to-right and right-to-left \textsc{lstm}s generate each a hidden vector of size 400.
\end{itemize}

\end{document}